\documentclass[sigconf]{acmart}

\usepackage{colortbl}
\usepackage{wrapfig}
\usepackage{booktabs}
\usepackage{graphicx}
\usepackage{enumitem}
\usepackage{algorithm}
\usepackage{listings}
\usepackage{pifont}
\usepackage{caption}
\usepackage{needspace}
\usepackage{subcaption}
\usepackage{multirow}
\usepackage{algorithm}
\usepackage{algpseudocode}
\usepackage{afterpage}
\usepackage{tcolorbox}

\definecolor{RowHighlight}{gray}{0.9}
\usepackage{hyperref}
\providecommand{\todo}[1]{}

\copyrightyear{2026}
\acmYear{2026}
\setcopyright{cc}
\setcctype{by}
\acmConference[KDD '26]{Proceedings of the 32nd ACM SIGKDD Conference on Knowledge Discovery and Data Mining V.2}{August 09--13, 2026}{Jeju Island, Republic of Korea}
\acmBooktitle{Proceedings of the 32nd ACM SIGKDD Conference on Knowledge Discovery and Data Mining V.2 (KDD '26), August 09--13, 2026, Jeju Island, Republic of Korea}
\acmDOI{10.1145/3770855.3817895}
\acmISBN{979-8-4007-2259-2/2026/08}

\usepackage{xspace}
\newcommand{\methodname}{\text{GRIP}\xspace}

\begin{document}

\title{\methodname: In-Parameter Graph Reasoning through Fine-Tuning Large Language Models}

%%
%% Author block. The names, ORCIDs, emails, institutions, departments,
%% and locations below MUST stay identical to the values entered on the
%% ACM RightsReview form (character-by-character, including punctuation,
%% capitalisation, and diacritics). Any discrepancy voids the form.

\author{Jiarui Feng}
\orcid{0000-0002-3409-6819}
\email{feng.jiarui@wustl.edu}
\affiliation{%
  \institution{Washington University in Saint Louis}
  \department{Computer Science and Engineering Department}
  \city{Saint Louis}
  \state{MO}
  \country{United States}
}

\author{Donghong Cai}
\orcid{0009-0005-5790-7505}
\email{cai.d@wustl.edu}
\affiliation{%
  \institution{Washington University in Saint Louis}
  \department{Computer Science and Engineering Department}
  \city{Saint Louis}
  \state{MO}
  \country{United States}
}

\author{Yixin Chen}
\orcid{0000-0002-3704-4432}
\email{chen@cse.wustl.edu}
\affiliation{%
  \institution{Washington University in Saint Louis}
  \department{Computer Science and Engineering Department}
  \city{Saint Louis}
  \state{MO}
  \country{United States}
}

\author{Muhan Zhang}
\authornote{Corresponding author.}
\orcid{0000-0002-7680-6401}
\email{muhan@pku.edu.cn}
\affiliation{%
  \institution{Peking University}
  \department{Institute for Artificial Intelligence}
  \city{Beijing}
  \country{China}
}

\renewcommand{\shortauthors}{Jiarui Feng, Donghong Cai, Yixin Chen, and Muhan Zhang}


\begin{abstract}
Large Language Models (LLMs) have demonstrated remarkable capabilities in modeling sequential textual data and generalizing across diverse tasks. However, effectively adapting LLMs to structural data, such as knowledge graphs or web graphs, remains a fundamental challenge. Some approaches adopt complex strategies to convert graphs into text sequences, resulting in significant token overhead and rendering them impractical for large-scale graphs. Others introduce additional modules to encode graphs into fixed-size token representations for LLMs. However, these methods typically require large-scale post-training on graph-text corpus and complex alignment procedures, yet often yield sub-optimal results due to poor modality alignment. In this work, we propose \methodname. Instead of relying on heavy graph serialization or specialized graph encoding modules, \methodname directly internalizes complex relational knowledge from graphs into the parameters of LLM through carefully designed fine-tuning tasks. The acquired structural knowledge is compactly stored in lightweight LoRA modules, enabling the fine-tuned LLM to perform a wide range of tasks over the internalized graph \textbf{without requiring access to the original graph as context} at inference time. Extensive experiments validate our approach. For graphs that cannot fit within the LLM’s context window, \methodname consistently outperforms LLM baselines by leveraging internalized graph knowledge, while for small-scale graphs, it achieves comparable performance with substantially lower inference cost.
\end{abstract}

\begin{CCSXML}
<ccs2012>
   <concept>
       <concept_id>10010147.10010257</concept_id>
       <concept_desc>Computing methodologies~Machine learning</concept_desc>
       <concept_significance>500</concept_significance>
       </concept>
 </ccs2012>
\end{CCSXML}

\ccsdesc[500]{Computing methodologies~Knowledge representation and reasoning}

\keywords{Graph Learning, Graph Algorithm, Test Time Adaptation, Large Language Models}


\maketitle

\section{Introduction}
In recent years, Large Language Models (LLMs) such as ChatGPT~\cite{chatgpt} and DeepSeek~\cite{deepseek} have revolutionized the field of artificial intelligence. Pre-trained on large-scale corpora of human knowledge using next-token prediction, these models have demonstrated remarkable generalization capabilities across a variety of downstream tasks, including math problem solving~\cite{GSM8K}, coding~\cite{deepseekcoder}, tool use~\cite{hugginggpt}, and knowledge-intensive applications~\cite{rag, paperqa}. However, to fully leverage the strong reasoning abilities of LLMs, tasks must first be formulated and expressed in human-readable textual formats. This conversion process is often complex and, in some cases, infeasible—particularly for tasks involving intricate data structures such as graphs. 

There has been extensive work on adapting LLMs for graph-related tasks, which can be broadly classified into two categories. The first class of approaches directly applies LLMs to graph data by converting graphs into text sequences~\cite{NLGraph, instructGLM, langGFM}. However, due to the complex structure of graphs, representing them as sequences is nontrivial and often results in \textbf{excessive token overhead or degraded inference performance}~\cite{instructGLM}, making this approach impractical for large graphs. In particular, for scenarios requiring \emph{reasoning over the same graph with multiple queries}—common in web-based retrieval or knowledge graph reasoning—the overhead scales linearly with the number of queries, since the same graph context must be combined with each query repeatedly. Moreover, for large-scale graphs, the size of the graph can easily exceed the context window of LLMs. A recent study has also highlighted the inherent limitation of sequential LLMs in solving graph tasks under node permutation~\cite{gnntaskplanner}. We provide a more detailed discussion in the next section.  

The second class of approaches integrates specialized modules, such as Graph Neural Networks (GNNs), to process graph data prior to interaction with the LLM. The resulting graph embeddings are then aligned with the LLM’s representation space via an additional adapter and alignment~\cite{gofa, graphgpt, llaga}. While effective in some cases, these methods typically require \textbf{careful model design and fine-tuning tasks to achieve alignment between language and graph modalities}. Furthermore, such frameworks typically require modifications to the model itself, which prevents the use of common acceleration techniques designed for standard LLMs and restricts their practical applicability.

% Inspired by recent advances in parameterized knowledge injection for LLM test-time adaptation~\cite{Temp-Lora, lift, parametricRAG}, we investigate the potential of embedding graph information directly into LLM parameters using Parameter-Efficient Fine-Tuning (PEFT) techniques such as LoRA~\cite{lora}. 

In this work, we explore the potential of embedding graph information directly into LLM parameters. To this end, we introduce Graph Reasoning In-Parameterization (\methodname), a unified framework illustrated in Figure~\ref{fig:overview}. In \methodname, we generate diverse fine-tuning tasks to guide LLMs in both memorizing graph contexts and leveraging this knowledge for downstream tasks. Given an input graph, we design context memorization and summarization tasks to inject graph knowledge into the LoRA parameters. We then construct various QA tasks on the graph to enable the LLM to effectively retrieve and apply the memorized knowledge to answer queries. Through fine-tuning, the LLM internalizes the structural information of the graph and encodes it into the LoRA parameters. At inference time, since the graph context is already embedded in the parameters, the model can perform a wide range of tasks related to the injected graph without requiring explicit graph context, simply by applying the trained LoRA. \methodname offers several key advantages:

\begin{itemize}
    \item \textbf{Simplicity.} It eliminates the need for specialized graph modules or graph-to-sequence conversions during inference.
    \item \textbf{Generalization ability.} The framework of \methodname\ is largely agnostic to downstream tasks and applies broadly across different graphs by training a graph-specific LoRA (rather than transferring a single LoRA across unseen graphs).
    \item \textbf{Efficiency.} After training, it can perform various tasks on the trained graph without requiring access to the original graph data as context, which significantly reduces token cost and improves inference efficiency.
    \item \textbf{Reusability.} Once the LoRA is trained, it can be stored and reused for inference on the same graph at any time.
\end{itemize}

We conduct extensive experiments to validate the effectiveness of \methodname. For graph tasks where the graph size exceeds the context window of the LLM, \methodname\ consistently outperforms baseline LLMs with graph-to-sequence conversion. This improvement stems from its ability to internalize the entire graph into the LoRA parameters and leverage this knowledge during inference. For tasks where the graph fits within the LLM’s context window, \methodname\ achieves comparable performance to baselines without requiring explicit graph context, demonstrating its efficiency. Moreover, when provided with the graph context, \methodname\ further improves performance and surpasses the baselines, highlighting its potential as a test-time adaptation method for graph tasks. We also provide a detailed discussion of the limitations and practical application scenarios of \methodname. Overall, the comprehensive results and analysis demonstrate the strong potential of \methodname\ to enhance the test-time adaptation of LLMs for structured graph data.

\section{Related Works}
\label{sec:related_works}
The success of the foundational language models inspired many works to adapt them to the graph domains and design a foundational model on the graph domain. Typically, existing methods can be divided into two categories: GNN-LLM-based and pure LLM-based. 

\textbf{GNN-LLM based methods.} This line of work typically focuses on designing specialized modules for graph data processing, leveraging the capabilities of LLMs to enhance the model’s generalization ability. These methods can be broadly categorized into two groups: LLM-as-Enhancer and LLM-as-Predictor. LLM-as-Enhancer approaches utilize LLMs to unify the input space, enabling cross-domain inference across various types of graph data. For example, OFA~\cite{ofa} employs LLMs to standardize input features from different datasets, transforming multiple graph classification tasks into a unified binary classification format. TAPE~\cite{tape} uses LLMs to generate question-answer pairs and explanations as enriched node features to improve learning. LLM-as-Predictor methods, on the other hand, aim to align embeddings learned from graph models with the representation space of LLMs. For instance, GraphGPT~\cite{graphgpt} fine-tunes a projection module to align embeddings between a pretrained GNN and an LLM. LLaGA~\citep{llaga} introduces a creative template-based approach that represents subgraphs using pooled node embeddings for LLM input. Inspired by Q-former~\citep{blip2}, GraphTranslator~\citep{graphtranslator} aligns node and text tokens by connecting pretrained GNN and LLM representations. UniGraph~\citep{unigraph} pretrains a GNN using masked word prediction and then learns a projection function to map graph embeddings into the language space, enabling zero-shot inference. However, these methods often require a carefully designed alignment module and task formulations to achieve successful alignment. Furthermore, since many of these methods require careful modifications to the original LLM, common acceleration techniques designed for standard LLMs become inapplicable, which limits their practical usage. For example, GL-Fusion~\cite{glfusion} adapts graph data into an LLM by modifying the original attention module of the transformer to incorporate graph bias. However, such modifications prevent the use of many modern inference accelerations designed for attention modules, such as vLLM~\cite{vllm}.

\textbf{LLM-based methods.} Many researchers have also explored the potential of directly using LLMs for graph reasoning. For example, NLGraph~\cite{NLGraph} represents graphs as sequences and evaluates this approach on various structural tasks. InstructGLM~\cite{instructGLM} further investigates alternative strategies for describing graph data in textual form. LangGFM~\cite{langGFM} directly fine-tunes LLMs on graph sequences and achieves impressive results across several graph benchmarks. Similarly, Li et al.~\cite{li2025largelanguagemodelsincontext} interpret the GNN process as a form of Retrieval-Augmented Generation (RAG) and design specific patterns to represent input graphs as text sequences that simulate GNN computations. PromptGFM~\cite{promptgfm} also follows this idea by directly simulating the message-passing process of GNNs using LLMs. However, these methods inherently rely on converting graphs into sequences—a nontrivial and often challenging task. A common strategy is to represent the graph using its edge list as input to the LLM. Yet, for graphs with high node degrees or long node feature representations, this approach leads to substantial token overhead, since node information must be redundantly repeated for each connected edge. Moreover, recent work~\cite{gnntaskplanner} has highlighted that LLMs are fundamentally limited in performing graph reasoning tasks due to their sensitivity to the ordering of nodes and edges; such permutations can significantly impact downstream performance. 

\textbf{Parameterized continual learning for LLMs.} Recently, the concept of continual learning for LLMs has gained significant attention due to the growing need to adapt LLMs to the latest knowledge or domain-specific knowledge during test time. Among various approaches, parameterized continual learning has emerged as a promising direction, offering efficient inference by directly injecting new knowledge into model parameters or PEFT adapters. For example, Parametric RAG~\cite{parametricRAG} fine-tunes a unique LoRA adapter for each document, allowing the system to retrieve and apply the corresponding adapter at inference time based on the user query, rather than retrieving raw text chunks. DyPRAG~\cite{dyprag} and text-to-lora~\cite{text_to_lora} further advance this idea by training a meta-network to dynamically predict LoRA weights from input documents. Other methods explore the potential to merge multiple LoRA for knowledge combination and enhancement, including~\cite{lorahub, knots}. Finally, LIFT~\cite{lift} and Temp-Lora~\cite{Temp-Lora} explore the potential of fine-tuning long-context inputs into PEFT parameters, showing the great potential on parameterized continual learning on long context. However, to the best of our knowledge, no existing work has extended these ideas to the graph domain or analyzed its potential on graph-structured data.

\begin{table*}[h]
\centering
\caption{Ablation study on graph-to-sequence methods on Qwen2.5-7B.}
\label{tab:ablationp_gts}

\begin{tabular}{@{}l c | c c | c c @{}}
\toprule
\multirow{2}{*}{Method} &
\multirow{2}{*}{Theoretical avg. token} &
\multicolumn{2}{c|}{Scene graph} &
\multicolumn{2}{c}{Clegr-reasoning} \\
\cmidrule(lr){3-4}\cmidrule(lr){5-6}
 & & Practical avg. token & Accuracy & Practical avg. token & Accuracy \\
\midrule
edge with index & $nt_n + ndt_e$   & 2296.28 & 46.60 & 3938.22 & 29.37 \\
edge list       & $2ndt_n + ndt_e$ & 5295.33 & 67.46 & 6338.80 & 33.48 \\
\bottomrule
\vspace{-5pt}
\end{tabular}
\end{table*}

\section{The limitation of graph-to-sequence based inference}
\label{app:graph_to_seq}
We begin by discussing the limitations of the standard LLM inference pipeline for graph data. To enable an LLM to directly address graph tasks, a crucial step is to represent the graph as a text sequence. Several graph-to-sequence methods have been proposed; due to space constraints, we focus on two representative approaches. Let the average number of tokens for node and edge features be denoted by $t_n$ and $t_e$, respectively, and let the average node in-degree and out-degree be $d$. The minimum total number of tokens required to represent a graph with $n$ nodes is thus $n t_n + n d t_e$. The two standard graph-to-sequence methods are (1) edge with index and (2) edge list.

\begin{tcolorbox}[colback=cyan!2!white, colframe=blue!30!white, boxsep=0mm,left=2.5mm,right=2.5mm]
\textbf{Edge with index:}\\
Node list: \\
1. the man; 2. the tree; 3. the bird; 4. the chair;

Edge list: \\ 
1 is on the left of 2; 1 is sitting on 4; 3 is above 1; 3 is above 2;

\end{tcolorbox}

\textbf{Edge with index.} For the edge with index method, the graph representation begins by listing all nodes as a sequence, assigning each node a unique ID. For example, each node is represented as "ID: node feature." Each edge is then expressed as a text sequence of the form "SRC\_ID, edge feature, TGT\_ID," where SRC\_ID and TGT\_ID denote the source and target node IDs, respectively. The final sequence representation of the graph is constructed by concatenating all node and edge descriptions. An illustrative example is shown above. Since node IDs are typically simple tokens whose cost can be ignored, the total number of tokens required under the edge with index method is $n t_n + n d t_e$, which matches the theoretical minimum token count. However, because each edge contains only node IDs rather than full node features, the LLM must retrieve the corresponding features via these IDs, introducing additional reasoning complexity—particularly for large graphs.

\begin{tcolorbox}[colback=cyan!2!white, colframe=blue!30!white, boxsep=0mm,left=2.5mm,right=2.5mm]
\textbf{Edge list:} \\
the man is on the left of the tree; the man is sitting on the chair; the bird is above the man; the bird is above the tree.
\end{tcolorbox}

\textbf{Edge list.} Another approach is to directly provide the LLM with the full edge list. In this case, each edge is represented as "source node feature, edge feature, target node feature," and there is no need to separately describe all nodes. An illustrative example is shown above. This method seems more advantageous than the previous one, as the LLM can directly access the contextual information of each edge without resolving node IDs. However, because the feature representation of each node is repeated for every connected edge, the node information is redundantly included as many times as the node’s degree. Consequently, the total number of tokens required to represent the graph becomes $2ndt_n + ndt_e$ (count both in-degree and out-degree), which can be substantially larger than $nt_n + ndt_e$ when the average degree $d$ is large (as is common in protein–protein interaction or web networks) or when the node feature size $t_n$ is large (as often observed in citation networks).

Different graph-to-sequence methods often result in different performance on LLM inference. In Table~\ref{tab:ablationp_gts}, we compare the performance of the LLM using both edge with index and edge list methods on two graph tasks with Qwen2.5-7b~\cite{qwen2}. Although the edge with index method yields a significantly smaller average number of tokens compared to the edge list method, it also leads to degraded downstream performance. Note that other advanced methods for converting a graph to a sequence, such as PromptGFM~\cite{promptgfm} or Graph as RAG~\cite{li2025largelanguagemodelsincontext}, have also been proposed. However, these approaches typically incur even higher token costs than the edge list, and are therefore not considered in this work. Another limitation of graph-to-sequence conversion is that it inevitably destroys the permutation-invariant property of graphs, making the performance of LLMs highly sensitive to the ordering of nodes and edges~\cite{gnntaskplanner}. Finally, in many practical scenarios, graphs are stored as static databases with a large number of diverse queries or tasks posed over them. When using graph-to-sequence methods, the entire graph context must be appended to each query, which leads to the token overhead on the graph-to-sequence method to grow linearly with the number of queries. Moreover, many real-world graphs are too large to fit within the context window of an LLM, and subgraph sampling is required, which further complicates inference.

Given these limitations, a natural question arises: how can LLMs be applied to graph tasks more effectively? We begin by noticing that LLMs already encode substantial knowledge in their parameters during pretraining and post-training, enabling them to solve many tasks without additional context. This observation motivates the idea of directly injecting graph information into the model parameters, allowing the LLM to perform inference on graphs by leveraging parameterized knowledge.

\begin{figure*}
    \includegraphics[width=0.98\textwidth, trim=0.5cm 7cm 0.5cm 3cm, clip]{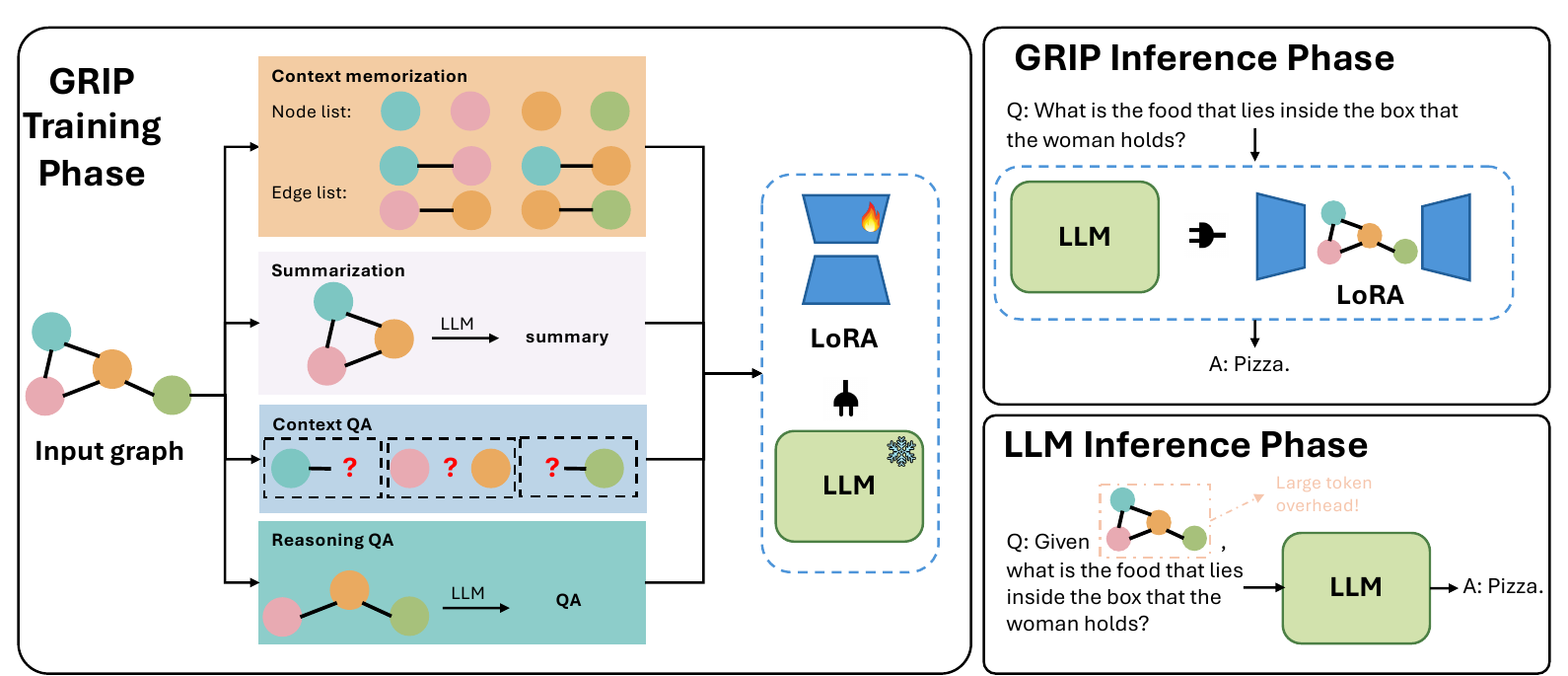}
    \caption{Overview of \methodname. During the fine-tuning phase, we design a variety of tasks to inject graph context into the LoRA parameters and explicitly instruct the model to utilize this context for solving downstream tasks. In the inference phase, \methodname\ can directly answer user queries without requiring explicit graph context. In contrast, standard LLM-based inference over graphs relies on providing explicit graph context, which introduces significant token overhead.}
    \label{fig:overview}
\end{figure*}

\section{Methods}
To this end, we propose \methodname. The overall framework of \methodname is illustrated in Figure~\ref{fig:overview}. In brief, \methodname injects graph information into the model parameters by generating multiple tasks on the graph with a unified framework and fine-tuning the LLM using LoRA~\cite{lora}. The training objectives are organized into two complementary groups: \emph{memorization} tasks (context memorization and summarization), which write graph facts and higher-order structure into the LoRA, and \emph{question answering} tasks (context QA and reasoning QA), which teach the model to retrieve and reason over the parameterized knowledge in a way the LLM can natively use at inference time. We discuss each component of \methodname in the following.

\subsection{Preliminaries}
Denote a text input by $\mathbf{x} = (x_1, \ldots, x_L)$ with length $L$. In this paper, we consider standard language modeling task, which can be formalized as $\mathcal{L}_{LM}(\mathbf{x}; \theta) = -\sum_{i=1}^{L} \log \mathbb{P}(x_i \mid \mathbf{x}_{0:i-1}; \theta)$. Low-Rank Adaptation (LoRA) is a widely used technique for efficiently fine-tuning LLMs. Instead of directly fine-tuning the original LLM parameter matrix $W \in \mathbb{R}^{d \times d}$, where $d$ is the hidden dimension of an LLM layer, LoRA learns two matrices $A \in \mathbb{R}^{d \times r}$ and $B \in \mathbb{R}^{r \times d}$, where $r \ll d$. The update to the original weight is given by $\Delta W = A \cdot B$, and the final fine-tuned parameter becomes $\hat{W} = W + \Delta W$.

In this work, we primarily focus on Text-Attributed Graphs (TAGs). A TAG can be represented as $G = (V, E, X_V, X_E)$, where $V$ and $E$ are the sets of nodes and edges, respectively. Each edge $e \in E$ is represented by a triplet $e = (s, r, t)$, where $s$ is the source node, $r$ is the relation, and $t$ is the target node. Each node $v \in V$ and each edge $e \in E$ is associated with a textual description $\mathbf{x}_v \in X_V$ or $\mathbf{x}_e \in X_E$, respectively.

\subsection{Graph context memorization}
The overall principle of \methodname is to internalize graph context into the parameters of a LoRA adapter.  This requires the model to first memorize the relevant graph information. A straightforward approach is to apply existing graph-to-sequence methods to convert the entire graph into a long sequence and fine-tune the LoRA adapter to memorize it. However, this approach inherits the limitations discussed earlier, including substantial token overhead and sensitivity to node ordering permutations. We note that, while converting the entire graph into a sequence is required for accurate in-context inference, this is unnecessary for fine-tuning. Instead, in \methodname, we decompose the graph into smaller, independent components and encode them separately by fine-tuning the model to memorize individual features of nodes and edges directly. Formally, the task can be defined as:
\begin{equation}
\mathcal{L}_{context}=\sum_{v\in V}\mathcal{L}_{LM}(\mathbf{x}_v;\theta) + \sum_{e\in E}\mathcal{L}_{LM}(\mathbf{x}_e;\theta).
\end{equation}

This approach also avoids the node permutation issue and prevents the model from memorizing a node or edge in the condition of others (the inductive bias introduced by sequential modeling). 

However, independently fine-tuning the model on plain node and edge features may be insufficient for LLMs to memorize them. Moreover, it does not allow the model to capture the high-order structural information of a graph. To address this limitation, \methodname\ introduces the summarization task. Specifically, given a graph input, we randomly sample $N_s$ nodes, $N_s$ edges, and $N_s$ subgraphs. Each subgraph is rooted at a randomly selected node. The size of each subgraph can be controlled by the sampling strategy to avoid excessive token costs. We then prompt the LLM to generate a summary that captures the key information within each node/edge/subgraph (the detailed prompt design is provided in Appendix~\ref{app:prompts}). These summaries contain rich local and high-order structural information, and their union captures the entire graph information in a different format. To further enrich the tasks, we additionally prompt the LLM to rephrase each summary into different formats. Overall, it results in $6N_s$ different summaries. Denote each generated summary as $\mathbf{s}_i$. After obtaining all summaries, we fine-tune the LoRA adapter to memorize them by:
\begin{equation}
\mathcal{L}_{summary}=\sum_{i=1}^{6N_s}\mathcal{L}_{LM}(\mathbf{s}_i;\theta).
\end{equation}

By fine-tuning the LoRA to memorize these summaries, the model directly captures both local and higher-order structural and relational information from the graph data. It is worth noting that the summarization and rephrasing tasks also function as a form of knowledge augmentation, a strategy shown to be highly effective for knowledge storage and retrieval in LLMs~\cite{allen2023physics}.

\subsection{Enabling in-parameter graph reasoning through question answering}
So far, we have focused on enabling the model to memorize graph context. However, with only these tasks, the model still lacks the ability to effectively leverage memorized knowledge to solve downstream tasks, which require both retrieval and reasoning over graph information. To achieve this goal, we design two types of question answering (QA) tasks: context QA and reasoning QA. Next, we describe each QA task in detail.

\textbf{Context QA task.} The context QA task is designed to instruct the model on how to retrieve memorized knowledge. We divide the context QA task into node-level and edge-level. For the node-level task, we randomly sample $N_c$ nodes from the graph. For each node, we prompt the LLM to randomly mask one attribute of its feature and generate a QA pair to ask about the masked attribute. For the edge-level task, we randomly sample $N_c$ edges from the graph. For each edge $e = (s, r, t)$, we randomly mask one of $s$, $r$, or $t$ and ask the model to predict the missing component given the remaining two. For example, ``What is the relationship between $s$ and $t$?'' or ``Which node has the relationship of $r$ to node $t$?'' This results in a total $2N_c$ QA pairs. The detailed prompt and question templates are shown in Appendix~\ref{app:prompts} and Appendix~\ref{app:implementation_setting}. By training the model to answer such questions, we encourage the model to strengthen its ability to retrieve the encoded graph given the query.

\textbf{Reasoning QA task.} The reasoning QA task focuses on enhancing the LLM’s ability to perform reasoning over the entire graph. To achieve this, we first sample $N_r$ subgraphs, each rooted at a randomly selected node. We then prompt the LLM to generate reasoning questions and corresponding answers that can be answered by leveraging the provided subgraph. To diversify the types of questions, we design four distinct prompts, encouraging the LLM to generate questions from different reasoning categories. Specifically, the questions are divided into four types: multi-hop, global, binary, and $k$-shot. In brief, multi-hop questions test the model’s ability to reason over multiple pieces of information in the graph; global questions require reasoning over the entire sampled subgraph; binary questions are formulated to be answered with either “yes” or “no”; and $k$-shot questions leverage sample QA pairs from the training set (if have) of the dataset to generate new questions with similar formats. The detailed prompt for each type can be found in Appendix~\ref{app:prompts}. For each subgraph, we randomly select one question type for QA task generation and generate two QA pairs on it. This results in $2N_r$ QA pairs in total.

Let the $i$-th question and answer be denoted by $\mathbf{q}_i$ and $\mathbf{a}_i$, respectively. The QA objective is formalized as:
\begin{equation}
    \mathcal{L}_{QA} = -\sum_{i=1}^{2N_c}log\mathbb{P}(\mathbf{a}_i|\mathbf{q}_i;\theta) -\sum_{j=1}^{2N_r}log\mathbb{P}(\mathbf{a}_j|\mathbf{q}_j;\theta).
\end{equation}

\subsection{Fine-tuning and inference of \methodname}
To facilitate effective fine-tuning, \methodname\ divides the fine-tuning into two stages. In the first stage, we only fine-tune the LoRA on graph context memorization. The overall loss for the first stage is:
\begin{equation}
    \mathcal{L}_{stage1} = \mathcal{L}_{context} + \mathcal{L}_{summary}.
\end{equation}
In the second stage, we fine-tune the LoRA on question answering tasks:
\begin{equation}
\mathcal{L}_{stage2} = \mathcal{L}_{QA}
\end{equation}

Other training details can be found in Appendix~\ref{app:implementation_setting}. 

During inference, the model already encodes the graph context within the LoRA parameters, allowing it to directly solve downstream tasks without requiring the graph as explicit input. Let the downstream task be denoted by $\mathbf{t}$ and the graph context by $\mathbf{g}$. The inference process of \methodname\ is simply $\mathbf{a} = \text{\methodname}(\mathbf{t})$, whereas standard LLM inference requires $\mathbf{a} = \text{LLM}(\text{concat}(\mathbf{g}, \mathbf{t}))$. Since $\mathbf{g}$ typically forms a long sequence—especially for large-scale graphs—standard LLM inference demands a significantly larger context window and incurs higher computational cost compared to \methodname.

\subsection{Additional discussion on \methodname}
In this section, we provide additional discussion on \methodname.

\textbf{The training cost of \methodname.} First, although \methodname\ is more efficient than standard LLMs with graph-to-sequence conversion at inference time, it requires fine-tuning on carefully designed tasks, which introduces additional computational overhead compared to direct inference. For small-scale graphs, this fine-tuning cost may outweigh the inference savings. However, the practicality of \methodname\ becomes evident for large-scale graphs, particularly in scenarios involving frequent and repeated queries over the same graph, where the one-time fine-tuning cost can be amortized and leads to substantial overall efficiency gains.

\textbf{Comparison with RAG-based methods.} RAG-based methods are widely used for graph-related tasks, especially for large graphs, but they target a different goal from \methodname. RAG-based methods retrieve question-relevant subgraphs at inference time, whereas \methodname internalizes graph knowledge directly into the LLM parameters. Consequently, RAG-based performance heavily depends on retrieval quality and often requires increasingly complex retrieval and indexing designs~\cite{g_retriever, gfmrag}, while also introducing additional inference latency that typically grows at least linearly with graph size and accumulates under repeated queries. In contrast, \methodname only requires standard LoRA-based fine-tuning and incurs no additional overhead at inference time beyond a single LLM forward pass. Moreover, \methodname is complementary to RAG-based methods and can be naturally integrated to enhance the answering stage by providing implicit global graph knowledge alongside retrieved evidence.

\textbf{Dealing with multiple graphs or dynamic graphs.} \methodname is primarily designed for a single graph, where each graph is stored in an independent LoRA module. Extending \methodname to multiple graphs is nevertheless straightforward, as multiple graphs can be treated as a single graph with multiple connected components and jointly used for fine-tuning. In addition, techniques such as LoRA merging~\cite{lorahub} can be leveraged to directly combine knowledge learned from multiple LoRA modules. We leave a systematic study of multi-graph training and LoRA composition to future work. Regarding dynamic graphs, \methodname naturally supports graph growth through standard continual fine-tuning on newly added or updated subgraphs, which will be further discussed in the experimental section. Nevertheless, we acknowledge that \methodname may be less suitable for graphs that undergo frequent and large-scale changes. This limitation, however, is not specific to \methodname, but reflects a more general challenge of maintaining up-to-date knowledge in large foundation models, which is beyond the scope of this work.

\section{Experiments}
In this section, we conduct various experiments to validate the effectiveness of \methodname. Specifically, we aim to answer the following questions: \\
\textbf{Q1}: How well does \methodname\ compare to standard LLMs and existing graph-based models on different graph tasks? \\
\textbf{Q2}: How efficient is \methodname\ compared to standard graph-to-sequence methods? \\
\textbf{Q3}: How effective are the designed fine-tuning tasks in \methodname? \\
\textbf{Q4}: What are the key hyperparameter settings in \methodname? \\
\textbf{Q5}: Can \methodname\ deal with growing graphs? \\
More details about the implementation and experimental setting can be found in Appendix~\ref{app:implementation} and Appendix~\ref{app:experiment}, respectively. 

\begin{table}[t]
\centering
\caption{Performance on the knowledge graphs (IT: input token; OOC: out of context window).}
\label{tab:kg}
% \vspace{-5pt}
% \fontsize{8}{9}\selectfont
\setlength{\tabcolsep}{2.5pt}
\begin{tabular}{c|cc|cc}
\toprule
Model  & FB15K237 $\uparrow$ & Avg \#IT $\downarrow$ & WN18RR $\uparrow$ & Avg \#IT $\downarrow$\\
\midrule
\midrule
OFA & 70.84 & 1304.90 & 30.96 & 1036.48 \\
GOFA & 80.69 & 1304.90 & 32.89 & 1036.48 \\
\midrule
Qwen-7b & 79.65 & 155.74 & 14.74 & 156.01 \\
Qwen-7b$_{\text{context}}$ & 79.19 & 1330.56 & 41.19 & 1624.53 \\
\textbf{\methodname-qwen-7b} & \textbf{88.75} & 155.74 & \textbf{64.30} & 156.01 \\
\midrule
Llama-8b & 56.66 & 152.75 & 28.08 & 155.70\\
Llama-8b$_{\text{context}}$ & 44.68 & 1304.90 & 53.86 & 1036.48\\
\textbf{\methodname-llama-8b} & 77.79 & 152.75 & 63.26 & 155.70 \\

\midrule
Full-context & OOC & $>$10M & OOC & $>$4M \\
\bottomrule
\end{tabular}
\vspace{-5pt}

\end{table}

\subsection{Setup}

\textbf{Dataset.} We evaluate \methodname\ on two categories of graph tasks. The first category is the traditional knowledge graph completion task. We include FB15K237~\cite{fb15k237}, WN18RR~\cite{wn18rr}, CoDEx-Medium~\cite{codex}, and NELL23K~\cite{nell23k}. The task for all these datasets is to predict missing relations between entities. The second category consists of advanced reasoning tasks on graphs. We include Scene Graph~\cite{g_retriever} and CLEGR-Reasoning~\cite{clegr}. Scene Graph is a dataset in which each graph corresponds to an image, and the task is to answer various questions related to the image. CLEGR-Reasoning contains multiple synthetic subway graphs, and the task is to answer reasoning questions defined over these graphs. For these two multi-graph datasets, each graph is internalised into its own dedicated LoRA. Additional details for each dataset are provided in Appendix~\ref{app:data_stats}.

\textbf{Model and baselines.} To provide a comprehensive evaluation, we adopt both Qwen2.5-7b~\cite{qwen2} and Llama3.1-8b~\cite{llama3} as the base LLM for \methodname. For all fine-tuning task generation, we use Qwen2.5-7b~\cite{qwen2}. To ensure a fair comparison, we include baselines from the two categories discussed in Section~\ref{sec:related_works}: the LLM baselines (Qwen2.5-7b and Llama3.1-8b with edge-list context) represent graph-to-sequence methods, while OFA~\cite{ofa} and GOFA~\cite{gofa} represent graph-to-embedding methods with zero-shot inference. LLM baselines are run by us; OFA/GOFA numbers are taken from the original papers.

\textbf{Metrics.} We use accuracy as the primary evaluation metric for all datasets. To ensure fairness, for the Scene Graph and CLEGR-Reasoning datasets we adopt the LLM-as-Judge approach~\cite{li2024llmsasjudgescomprehensivesurveyllmbased}. Specifically, we use the Qwen2.5-32B model as the judge to determine whether the model outputs are semantically equivalent to the ground-truth labels. For all knowledge graph datasets, accuracy is computed directly using exact match.

\begin{table}[t]
\centering
\caption{Performance on the knowledge graphs continued (IT: input token; OOC: out of context window).}
\label{tab:kg2}
% \vspace{-5pt}

% \fontsize{8}{9}\selectfont
\setlength{\tabcolsep}{3.0pt}
\begin{tabular}{c|cc|cc}
\toprule
Model  & CoDEx $\uparrow$ & Avg \#IT $\downarrow$ & NELL23K $\uparrow$ & Avg \#IT $\downarrow$\\
\midrule
\midrule
Qwen-7b & 83.09 & 124.21 & 77.38 & 160.80 \\
Qwen-7b$_{\text{context}}$ & 73.32 & 1055.29 & 81.38 & 590.37 \\
\textbf{\methodname-qwen-7b} & \textbf{94.42} & 124.21 & \textbf{87.74} & 160.80 \\
\midrule
Llama-8b & 71.03 & 124.61 & 20.05 & 160.33\\
Llama-8b$_{\text{context}}$ & 47.46 & 1044.44 & 53.33 & 586.15\\
\textbf{\methodname-llama-8b} & 91.58 & 124.61 & 77.40 & 160.33 \\
\midrule
Full-context & OOC & $>$5M & OOC & $>$0.5M \\
\bottomrule
\end{tabular}
\vspace{-5pt}

\end{table}

\subsection{Results on performance comparison}

In this section, we present the main experimental results. To provide a more holistic view, for LLM baseline models, we evaluate two variants. In models labeled with the subscript context, the LLM is provided with both the question and the relevant graph context (represented by the edge list). For the knowledge graph datasets, where the entire graph is too large to fit within the LLM’s context window, we sample a subgraph centered around the two target entities for relation prediction, which serves as a retrieval-augmented baseline. Additionally, we include a row labeled full-context to indicate the total number of tokens required to represent the complete knowledge graphs when converted into sequential text. For models without the subscript, only the question is provided as input—identical to the input used for \methodname. Performance on the knowledge graph datasets is reported in Table~\ref{tab:kg} and Table~\ref{tab:kg2}, while the results on the two reasoning datasets are shown in Table~\ref{tab:reasoning}. Note that for WN18RR, we observe high variance across runs; therefore, we report the average performance over five runs. For the other datasets, results are consistent across runs, so we report the single-run result.

To answer \textbf{Q1}, we first observe that \methodname\ consistently outperforms baseline LLMs without context across all datasets. This demonstrates the effectiveness of \methodname\ in injecting graph knowledge directly into the LoRA parameters, enabling task solving without explicit graph context. Next, we examine the results on knowledge graph datasets and compare \methodname\ to models that utilize explicit graph context. We find that \methodname\ achieves superior performance on nearly all datasets compared to both graph model and LLM model baselines (as indicated by bold scores). This improvement can be attributed to two factors. First, during fine-tuning, \methodname\ \textbf{injects knowledge from the entire graph into the model parameters}—a capability absent in baseline LLM methods, which rely on subgraph sampling to fit within the context window. As a result, \methodname\ leverages information from the entire graph to better solve downstream tasks. Second, these tasks evaluate the model on link prediction. Since \methodname\ incorporates similar link prediction tasks through context QA as part of its training objectives, they are effectively closer to in-domain evaluations. We also note that the context-based LLM baseline can occasionally underperform its no-context counterpart (e.g.\ Qwen-7b on CoDEx): the sampled subgraph around the queried entities is sometimes incomplete or distracting, which can mislead the model.
\begin{table}[t]
\centering
\vspace{-5pt}
\caption{Performance on the reasoning datasets (IT: input token; Scene: Scene Graph; Clegr: Clegr-reasoning).}
\label{tab:reasoning}

% \fontsize{9}{10}\selectfont
\setlength{\tabcolsep}{3.0pt}
\begin{tabular}{c|cc|cc}
\toprule
Model  & Scene $\uparrow$ & Avg \#IT $\downarrow$   & Clegr $\uparrow$ & Avg \#IT $\downarrow$   \\
\midrule
\midrule
GOFA & 34.06 & 4376.18 & - & -\\
\midrule
Qwen-7b & 3.54 & 67.94 & 15.81 & 106.41\\
Qwen-7b$_{\text{context}}$ & 67.46 & 5295.33 & \textbf{33.48} & 6338.8\\
\methodname-qwen-7b & 53.27 & 67.94 & 27.66 & 106.41 \\
\methodname-qwen-7b$_{\text{context}}$ & 68.71 & 5295.33  & 31.53 & 6338.8\\
\midrule
Llama-8b & 6.96 & 64.99 & 0.0 & 98.58\\
Llama-8b$_{\text{context}}$ & 67.06 & 4376.18 & 24.76 & 6213.93\\
\methodname-llama-8b & 53.27 & 64.99 & 27.14 & 98.58\\
\methodname-llama-8b$_{\text{context}}$ & \textbf{73.99} & 4376.18 & 32.75 & 6213.93\\
\bottomrule
\end{tabular}
\vspace{-5pt}

\end{table}

Next, we turn to the two reasoning tasks in Table~\ref{tab:reasoning}. These datasets correspond to a small-graph regime in which the full graph fits in the LLM’s context window, so context-based inference is naturally favoured; the goal here is therefore not for \methodname\ to dominate the full-context baseline, but to remain competitive at much lower inference cost and to further improve over the baseline when used as a test-time adaptation. Although \methodname\ does not surpass baseline LLMs with full context on these tasks, its performance remains comparable, with only a minimal gap. We would like to emphasize that baseline LLMs achieve these results by inputting the \textbf{full graph context}, while \methodname\ attains similar accuracy without any context, which is fundamentally different and indicates the strong potential of \methodname. To provide further evidence, we also evaluate \methodname\ under the same context as baseline LLMs on the reasoning tasks, and we observe that \methodname\ consistently outperforms the baselines. These results indicate that \textbf{\methodname\ can also serve as an effective test-time training method for enhancing the performance of LLMs on graph tasks.}

In summary, for graphs too large to fit into the context windows of LLMs, \methodname\ consistently outperforms baseline methods by encoding the entire graph into the model parameters. For graphs that fit within the context window, \methodname\ achieves comparable performance without any graph context and surpasses baselines when provided with the same context. These results collectively confirm the effectiveness of \methodname.

\subsection{Results on the efficiency}
\begin{figure}[htbp]
    \includegraphics[width=0.98\linewidth]{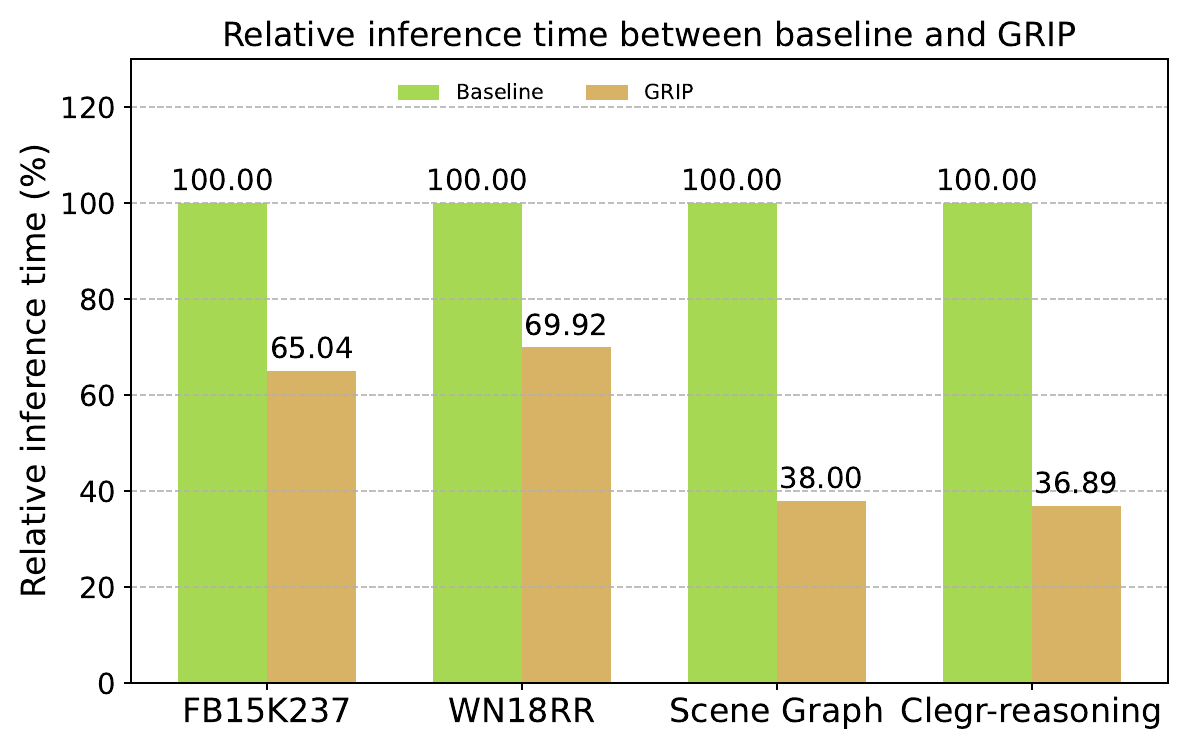}

    \caption{Relative inference time comparison between baseline and GRIP.}
    \label{fig:inference}
\end{figure}

\begin{table}[htbp]
\centering
\caption{Inference time (seconds) on test dataset with Qwen2.5-7b.}
% \vspace{-5pt}

\label{tab:inference_time}
% \fontsize{9}{10}\selectfont
% \setlength{\tabcolsep}{3.0pt}
\begin{tabular}{c|cc}
\toprule
dataset & Standard (with context) & \methodname \\
\midrule
\midrule
FB15K237 & 6316.71 & 4108.45  \\
WN18RR & 1157.69 & 809.47 \\
Scene Graph & 1083.69 & 411.81 \\
Clegr-reasoning & 17122.46 & 6316.52 \\
\bottomrule
\end{tabular}
\vspace{-5pt}

\end{table}

\begin{table}[t]
\centering
\setlength{\tabcolsep}{4pt}
\caption{Cost analysis of \methodname\ on FB15K237. Training time is one-time only; inference numbers are averaged per query.}
\label{tab:cost_fb15k237}
\begin{tabular}{lcccc}
\toprule
Method & Train (s) & Inf. (ms) & Avg.\ \#IT & Speedup \\
\midrule
Llama3.1-8B    & --     & 83.41 & 1326.27 & $1.00\times$ \\
\methodname\ (Llama3.1-8B) & 15{,}850 & 10.62 & 155.74  & $7.85\times$ \\
\bottomrule
\end{tabular}
\end{table}

To answer \textbf{Q2}, we provide the inference time and its comparison with baseline LLMs and \methodname on multiple datasets using Qwen2.5-7b, as shown in Table~\ref{tab:inference_time} and Figure~\ref{fig:inference}. Since \methodname does not require explicit access to the graph context during inference, it achieves substantially lower inference time compared to standard LLMs. Specifically, on knowledge graph datasets where baseline LLMs rely on subgraphs as context, \methodname reduces inference time by about 30\%. On reasoning datasets where baseline LLMs require the full graph as context, \methodname\ reduces inference time by more than 60\%. These results further demonstrate the efficiency of \methodname.  It is worth noting that \methodname does involve an additional fine-tuning phase. However, once fine-tuned, the resulting LoRA parameters can be stored and reused for any future inference tasks on the same graph. Consequently, the one-time fine-tuning cost becomes negligible compared to the cumulative inference cost, especially in scenarios such as GraphRAG or large-scale web graph applications, where inference is performed repeatedly. To make this concrete, Table~\ref{tab:cost_fb15k237} reports the wall-clock training and inference cost of \methodname\ versus the Llama3.1-8B baseline on FB15K237.

\subsection{Ablation studies on \methodname design}

To answer \textbf{Q3}, \textbf{Q4} and \textbf{Q5}, we conduct multiple ablation studies on \methodname. All experiments in this section are based on the Qwen2.5-7b model.

\textbf{Effects of fine-tuning task.} First, we evaluate the effectiveness of each fine-tuning task in \methodname\ by varying the numbers of $N_s$, $N_c$, and $N_r$ on both the Scene Graph and FB15K237 datasets. The results, shown in Table~\ref{tab:ablation_task_sg} and Table~\ref{tab:ablation_task_fb}, indicate that removing any task degrades overall performance, directly demonstrating the contribution of each task to \methodname. In particular, for the Scene Graph dataset, removing reasoning QA and summarization tasks results in the largest performance drop, while for FB15K237, the context QA contributes most significantly. This aligns with intuition: Scene Graph requires more complex reasoning, whereas FB15K237 centers on simple link prediction. These findings highlight the versatility of \methodname\ across different types of graph tasks.

\begin{table}[h]
\centering
\large
\caption{Effects of summarization and QA tasks on Scene Graph.}

\label{tab:ablation_task_sg}
\begin{tabular}{ccc|c}
\toprule
$N_s$ & $N_c$ &$N_r$ & Accuracy $\uparrow$  \\
\midrule
0 & 0 & 0 & 3.54 \\
\midrule
0 & 50 & 100 & 45.94\\
50 & 0 & 100 & 52.54 \\
50 & 50 & 0 & 45.74 \\
\midrule
50 & 50 & 100 & 53.27 \\
\bottomrule
\end{tabular}
\vspace{-5pt}

\end{table}

\begin{table}[h]
\centering
\large
\caption{Effects of summarization and QA tasks on FB15K237.}
% \vspace{-5pt}

\label{tab:ablation_task_fb}
\begin{tabular}{ccc|c}
\toprule
$N_s$ & $N_c$ &$N_r$ & Accuracy $\uparrow$  \\
\midrule
\midrule
 0 & 0 & 0 & 0.1 \\
\midrule
6000 & 0 & 2000 & 81.34 \\
0 & 8000 & 2000 & 84.09 \\
6000 & 8000 & 0 & 81.62 \\
6000 & 8000 & 2000 & 88.75 \\
\bottomrule
\end{tabular}
\vspace{-5pt}

\end{table}

\textbf{Effects of LoRA module.} Next, we examine the LoRA setting in \methodname. In our experiments, we apply LoRA only to the MLP components of the LLM. To validate this choice, we test applying LoRA to the attention modules and to all modules, with results shown in Table~\ref{tab:ablation_lora_p}. The results show that applying LoRA to the MLP modules is most effective. A likely reason is that MLP modules serve as knowledge storage and processing units, while attention modules focus on retrieval. In our setting, knowledge storage is more critical, and the retrieval ability of the base LLM is already sufficient.

\begin{table}[h]
\centering
\caption{Ablation study on LoRA adapter position.}
\label{tab:ablation_lora_p}
% \vspace{-5pt}

% \fontsize{8}{9}\selectfont
% \setlength{\tabcolsep}{3.0pt}
\begin{tabular}{c|ccc}
\toprule
Accuracy & ATT & ALL & MLP \\
\midrule
\midrule
Scene Graph & 43.10 & 50.36 & 53.27 \\

\bottomrule
\end{tabular}
\vspace{-5pt}

\end{table}

% \begin{table}[h]
% \centering
% \caption{Ablation study on LoRA adapter rank.}
% \label{tab:ablationp_lora_r}
% % \vspace{-5pt}

% % \fontsize{8}{9}\selectfont
% % \setlength{\tabcolsep}{3.0pt}
% \begin{tabular}{c|cccccc}
% \toprule
% Accuracy & r=1 & r=2 & r=4 & r=8 & r=16 & r=24 \\
% \midrule
% \midrule
% FB5K237 & 74.66 & 80.90 & 81.23 & 82.72 & 84.41 & 88.75 \\
% CoDEx & 90.58 & 91.65 & 94.42 & 93.37 & 92.58 & - \\
% NELL23K & 83.53 & 84.58 & 87.74 & 84.12 & 83.99 & - \\

% \bottomrule
% \end{tabular}
% \vspace{-5pt}

% \end{table}

\begin{figure}[t]

    \includegraphics[width=0.95\linewidth]{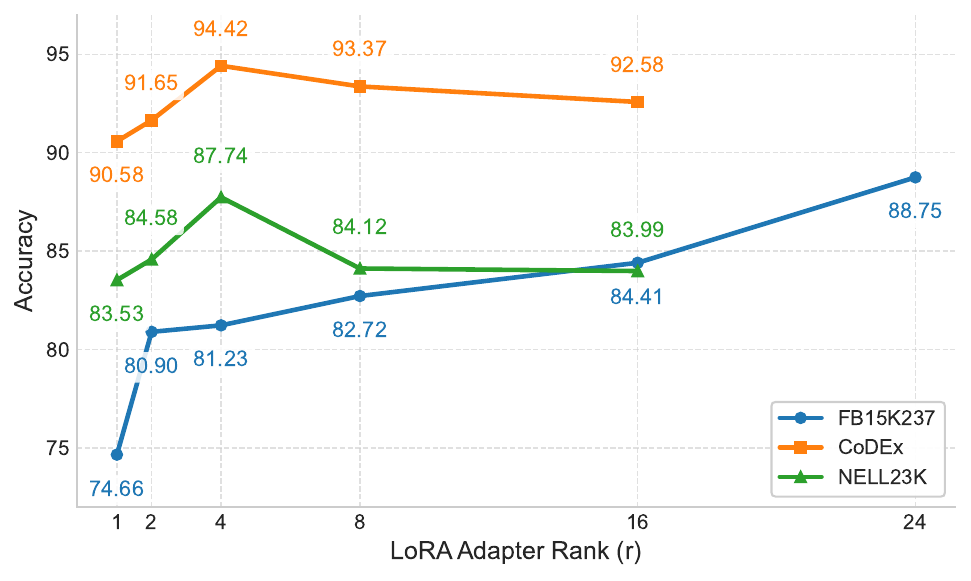}
    \vspace{-10pt}

    \caption{Ablation study on LoRA rank.}
    \label{fig:lora_rank}
    \vspace{-10pt}

\end{figure}

\textbf{Effects of LoRA rank. } We further study the effect of the LoRA rank $r$, and report the results in Figure~\ref{fig:lora_rank}. Several observations can be made. First, \methodname already achieves strong performance on large knowledge graphs even with a rank of $r=1$, indicating that the memorized graph knowledge is stored in a highly compressed form. Second, different graphs exhibit different optimal rank settings, and the preferred rank is positively correlated with graph size. In particular, \textsc{FB15k-237} requires a higher rank (up to $r=24$) to achieve the best performance, whereas for \textsc{CoDEx} and \textsc{NELL23K}, a small rank of $r=4$ is sufficient to reach near-optimal performance on downstream tasks. However, the best LoRA rank may depend not only on the size of the graph, but also on the downstream tasks. 

\textbf{Dealing with a growing graph.} Finally, we evaluate the performance of \methodname under periodically growing graphs. Specifically, we partition each knowledge graph into three subgraphs using a BFS-based strategy to ensure connectivity. Then, we train a single LoRA module in three consecutive stages with \methodname. At each stage, training tasks are generated only from the newly added subgraph. For a fair comparison, we sample one-third of the training tasks at each stage compared to the full setting, and evaluate the resulting model on the same full evaluation set. The results are reported in Table~\ref{tab:ablation_multistage}. Overall, \methodname maintains reasonable performance in the multi-stage setting, with accuracy generally improving as more graph parts are incorporated. However, the behavior varies across datasets. On \textsc{NELL23K}, performance is relatively weak in the first two stages and remains below the standard setting even after the third stage, whereas on \textsc{WN18RR}, the multi-stage setting slightly outperforms the standard setting. We attribute this variation mainly to the graph partition strategy, as the distribution of partial graphs may deviate from that of the original graph. In addition, training only on newly added subgraphs at each stage may induce forgetting of previously learned knowledge. We leave the design of more effective multi-stage training strategies for \methodname to future work.

\begin{table}[t]
\centering
\caption{Performance of \methodname\ on multi-stage setting. (Standard refers to a single-stage setting.)}
\label{tab:ablation_multistage}
% \vspace{-5pt}

% \fontsize{8}{9}\selectfont
% \setlength{\tabcolsep}{3.0pt}
\begin{tabular}{c|ccc|c}
\toprule
Accuracy & stage 1 & stage 2 & stage 3 & standard \\
\midrule
\midrule
NELL23K & 68.72 & 63.57 & 82.44 & 87.74 \\
WN18RR & 51.91 & 65.95 & 69.66 & 64.30 \\

\bottomrule
\end{tabular}

\end{table}
\textbf{Robustness across LLM backbones.} To verify that the benefit of \methodname\ is not tied to the specific base LLMs used above, we additionally train \methodname\ on Qwen3-8B and Qwen3-14B; as shown in Table~\ref{tab:qwen3_backbones}, \methodname\ remains the best variant on both WN18RR and FB15K237 across both model sizes.

\begin{table}[t]
\centering
\setlength{\tabcolsep}{4pt}
\caption{Accuracy of \methodname\ on Qwen3 backbones.}
\label{tab:qwen3_backbones}
\begin{tabular}{llccc}
\toprule
Backbone & Dataset & No Context & Context & \methodname \\
\midrule
\multirow{2}{*}{Qwen3-8B}  & WN18RR    & 25.50 & 60.79 & \textbf{68.48} \\
                            & FB15K237  & 77.03 & 74.06 & \textbf{91.01} \\
\midrule
\multirow{2}{*}{Qwen3-14B} & WN18RR    & 25.37 & 61.62 & \textbf{84.68} \\
                            & FB15K237  & 83.18 & 87.78 & \textbf{92.99} \\
\bottomrule
\end{tabular}
\end{table}

\section{Conclusion and Limitation}
In this work, we propose \methodname, a practical parameterised graph memory framework that adapts pre-trained LLMs to graph-structured data via parameter-efficient fine-tuning and in-parameter knowledge injection. \methodname provides a unified framework for generating fine-tuning tasks that inject graph context into LoRA parameters, enabling the model to reason over graphs without requiring explicit graph context or specialized graph-processing modules at inference time. We evaluate \methodname across multiple datasets and show that, for large-scale graphs exceeding the LLM’s context window, \methodname consistently outperforms subgraph-based baselines by effectively leveraging the memorized global graph. For graphs that fit within the context window, it achieves comparable performance without explicit context, highlighting efficiency. However, \methodname introduces additional training cost, which may exceed inference cost for small graphs, making it more suitable for large-scale settings. Moreover, \methodname is not well-suited for highly dynamic or frequently updated graphs, which reflects a broader limitation of large foundation models rather than a shortcoming specific to \methodname. We leave these directions to future work.

\begin{acks}
Jiarui Feng, Donghong Cai, and Yixin Chen are supported by a Washington University Transcend Initiative Grant and a DOE SFEE Moonshot grant. Muhan Zhang is supported by the National Natural Science Foundation of China (62550138, 62276003).
\end{acks}

\bibliographystyle{ACM-Reference-Format}
\bibliography{references}

@misc{chatgpt,
      title={GPT-4 Technical Report}, 
      author={OpenAI and et al.},
      year={2024},
      eprint={2303.08774},
      archivePrefix={arXiv},
      primaryClass={cs.CL},
      url={https://arxiv.org/abs/2303.08774}, 
}

@article{GSM8K,
  title={Training verifiers to solve math word problems},
  author={Cobbe, Karl and Kosaraju, Vineet and Bavarian, Mohammad and Chen, Mark and Jun, Heewoo and Kaiser, Lukasz and Plappert, Matthias and Tworek, Jerry and Hilton, Jacob and Nakano, Reiichiro and others},
  journal={arXiv preprint arXiv:2110.14168},
  year={2021}
}

@article{deepseek,
  title={Deepseek-v3 technical report},
  author={Liu, Aixin and Feng, Bei and Xue, Bing and Wang, Bingxuan and Wu, Bochao and Lu, Chengda and Zhao, Chenggang and Deng, Chengqi and Zhang, Chenyu and Ruan, Chong and others},
  journal={arXiv preprint arXiv:2412.19437},
  year={2024}
}

@article{deepseekcoder,
  title={DeepSeek-Coder: When the Large Language Model Meets Programming--The Rise of Code Intelligence},
  author={Guo, Daya and Zhu, Qihao and Yang, Dejian and Xie, Zhenda and Dong, Kai and Zhang, Wentao and Chen, Guanting and Bi, Xiao and Wu, Yu and Li, YK and others},
  journal={arXiv preprint arXiv:2401.14196},
  year={2024}
}

@article{hugginggpt,
  title={Hugginggpt: Solving ai tasks with chatgpt and its friends in hugging face},
  author={Shen, Yongliang and Song, Kaitao and Tan, Xu and Li, Dongsheng and Lu, Weiming and Zhuang, Yueting},
  journal={Advances in Neural Information Processing Systems},
  volume={36},
  pages={38154--38180},
  year={2023}
}

@article{rag,
  title={Retrieval-augmented generation for knowledge-intensive nlp tasks},
  author={Lewis, Patrick and Perez, Ethan and Piktus, Aleksandra and Petroni, Fabio and Karpukhin, Vladimir and Goyal, Naman and K{\"u}ttler, Heinrich and Lewis, Mike and Yih, Wen-tau and Rockt{\"a}schel, Tim and others},
  journal={Advances in neural information processing systems},
  volume={33},
  pages={9459--9474},
  year={2020}
}

@article{paperqa,
  title={Paperqa: Retrieval-augmented generative agent for scientific research},
  author={L{\'a}la, Jakub and O'Donoghue, Odhran and Shtedritski, Aleksandar and Cox, Sam and Rodriques, Samuel G and White, Andrew D},
  journal={arXiv preprint arXiv:2312.07559},
  year={2023}
}

@article{NLGraph,
  title={Can language models solve graph problems in natural language?},
  author={Wang, Heng and Feng, Shangbin and He, Tianxing and Tan, Zhaoxuan and Han, Xiaochuang and Tsvetkov, Yulia},
  journal={Advances in Neural Information Processing Systems},
  volume={36},
  pages={30840--30861},
  year={2023}
}

@inproceedings{instructGLM,
  title={Language is All a Graph Needs},
  author={Ye, Ruosong and Zhang, Caiqi and Wang, Runhui and Xu, Shuyuan and Zhang, Yongfeng},
  booktitle={EACL (Findings)},
  year={2024}
}

@article{langGFM,
  title={LangGFM: A Large Language Model Alone Can be a Powerful Graph Foundation Model},
  author={Lin, Tianqianjin and Yan, Pengwei and Song, Kaisong and Jiang, Zhuoren and Kang, Yangyang and Lin, Jun and Yuan, Weikang and Cao, Junjie and Sun, Changlong and Liu, Xiaozhong},
  journal={arXiv preprint arXiv:2410.14961},
  year={2024}
}

@inproceedings{gnntaskplanner,
  title={Can Graph Learning Improve Planning in LLM-based Agents?},
  author={Wu, Xixi and Shen, Yifei and Shan, Caihua and Song, Kaitao and Wang, Siwei and Zhang, Bohang and Feng, Jiarui and Cheng, Hong and Chen, Wei and Xiong, Yun and others},
  booktitle={The Thirty-eighth Annual Conference on Neural Information Processing Systems},
  year={2024}
}

@inproceedings{graphgpt,
  title={Graphgpt: Graph instruction tuning for large language models},
  author={Tang, Jiabin and Yang, Yuhao and Wei, Wei and Shi, Lei and Su, Lixin and Cheng, Suqi and Yin, Dawei and Huang, Chao},
  booktitle={Proceedings of the 47th International ACM SIGIR Conference on Research and Development in Information Retrieval},
  pages={491--500},
  year={2024}
}

@inproceedings{llaga,
  title={LLaGA: Large Language and Graph Assistant},
  author={Chen, Runjin and Zhao, Tong and Jaiswal, Ajay Kumar and Shah, Neil and Wang, Zhangyang},
  booktitle={International Conference on Machine Learning},
  pages={7809--7823},
  year={2024},
  organization={PMLR}
}

@inproceedings{gofa,
title={{GOFA}: A Generative One-For-All Model for Joint Graph Language Modeling},
author={Lecheng Kong and Jiarui Feng and Hao Liu and Chengsong Huang and Jiaxin Huang and Yixin Chen and Muhan Zhang},
booktitle={The Thirteenth International Conference on Learning Representations},
year={2025},
url={https://openreview.net/forum?id=mIjblC9hfm}
}

@article{allen2023physics,
  title={Physics of language models: Part 3.1, knowledge storage and extraction},
  author={Allen-Zhu, Zeyuan and Li, Yuanzhi},
  journal={arXiv preprint arXiv:2309.14316},
  year={2023}
}

@article{parametricRAG,
  title={Parametric Retrieval Augmented Generation},
  author={Su, Weihang and Tang, Yichen and Ai, Qingyao and Yan, Junxi and Wang, Changyue and Wang, Hongning and Ye, Ziyi and Zhou, Yujia and Liu, Yiqun},
  journal={arXiv preprint arXiv:2501.15915},
  year={2025}
}

@article{lift,
  title={LIFT: Improving Long Context Understanding Through Long Input Fine-Tuning},
  author={Mao, Yansheng and Li, Jiaqi and Meng, Fanxu and Xiong, Jing and Zheng, Zilong and Zhang, Muhan},
  journal={arXiv preprint arXiv:2412.13626},
  year={2024}
}

@article{Temp-Lora,
  title={With greater text comes greater necessity: Inference-time training helps long text generation},
  author={Wang, Yan and Ma, Dongyang and Cai, Deng},
  journal={arXiv preprint arXiv:2401.11504},
  year={2024}
}

@inproceedings{lora,
  title={LoRA: Low-Rank Adaptation of Large Language Models},
  author={Hu, Edward J and Wallis, Phillip and Allen-Zhu, Zeyuan and Li, Yuanzhi and Wang, Shean and Wang, Lu and Chen, Weizhu and others},
  booktitle={International Conference on Learning Representations},
  year={2021}

}

@article{g_retriever,
  title={G-retriever: Retrieval-augmented generation for textual graph understanding and question answering},
  author={He, Xiaoxin and Tian, Yijun and Sun, Yifei and Chawla, Nitesh and Laurent, Thomas and LeCun, Yann and Bresson, Xavier and Hooi, Bryan},
  journal={Advances in Neural Information Processing Systems},
  volume={37},
  pages={132876--132907},
  year={2024}
}

@article{fb15k237,
  title={Translating embeddings for modeling multi-relational data},
  author={Bordes, Antoine and Usunier, Nicolas and Garcia-Duran, Alberto and Weston, Jason and Yakhnenko, Oksana},
  journal={Advances in neural information processing systems},
  volume={26},
  year={2013}
}

@inproceedings{wn18rr,
	Author = {Dettmers, Tim and Pasquale, Minervini and Pontus, Stenetorp and Riedel, Sebastian},
	Booktitle = {Proceedings of the 32th AAAI Conference on Artificial Intelligence},
	Title = {Convolutional 2D Knowledge Graph Embeddings},
	Url = {https://arxiv.org/abs/1707.01476},
	Year = {2018},
        pages  = {1811--1818},
  	Month = {February}
}

@article{qwen2,
      title={Qwen2 Technical Report}, 
      author={An Yang and Baosong Yang and Binyuan Hui and Bo Zheng and Bowen Yu and Chang Zhou and Chengpeng Li and Chengyuan Li and Dayiheng Liu and Fei Huang and Guanting Dong and Haoran Wei and Huan Lin and Jialong Tang and Jialin Wang and Jian Yang and Jianhong Tu and Jianwei Zhang and Jianxin Ma and Jin Xu and Jingren Zhou and Jinze Bai and Jinzheng He and Junyang Lin and Kai Dang and Keming Lu and Keqin Chen and Kexin Yang and Mei Li and Mingfeng Xue and Na Ni and Pei Zhang and Peng Wang and Ru Peng and Rui Men and Ruize Gao and Runji Lin and Shijie Wang and Shuai Bai and Sinan Tan and Tianhang Zhu and Tianhao Li and Tianyu Liu and Wenbin Ge and Xiaodong Deng and Xiaohuan Zhou and Xingzhang Ren and Xinyu Zhang and Xipin Wei and Xuancheng Ren and Yang Fan and Yang Yao and Yichang Zhang and Yu Wan and Yunfei Chu and Yuqiong Liu and Zeyu Cui and Zhenru Zhang and Zhihao Fan},
      journal={arXiv preprint arXiv:2407.10671},
      year={2024}
}

@inproceedings{ofa,
  title={One For All: Towards Training One Graph Model For All Classification Tasks},
  author={Liu, Hao and Feng, Jiarui and Kong, Lecheng and Liang, Ningyue and Tao, Dacheng and Chen, Yixin and Zhang, Muhan},
  booktitle={The Twelfth International Conference on Learning Representations},
  year={2023}
}

@misc{li2024llmsasjudgescomprehensivesurveyllmbased,
      title={LLMs-as-Judges: A Comprehensive Survey on LLM-based Evaluation Methods}, 
      author={Haitao Li and Qian Dong and Junjie Chen and Huixue Su and Yujia Zhou and Qingyao Ai and Ziyi Ye and Yiqun Liu},
      year={2024},
      eprint={2412.05579},
      archivePrefix={arXiv},
      primaryClass={cs.CL},
      url={https://arxiv.org/abs/2412.05579}, 
}

@inproceedings{tape,
  title={Harnessing Explanations: LLM-to-LM Interpreter for Enhanced Text-Attributed Graph Representation Learning},
  author={He, Xiaoxin and Bresson, Xavier and Laurent, Thomas and Perold, Adam and LeCun, Yann and Hooi, Bryan},
  booktitle={The Twelfth International Conference on Learning Representations},
    year={2024},
}

@article{unigraph,
  title={UniGraph: Learning a Unified Cross-Domain Foundation Model for Text-Attributed Graphs},
  author={He, Yufei and Sui, Yuan and He, Xiaoxin and Hooi, Bryan},
  journal={arXiv preprint arXiv:2402.13630},
  year={2024}
}

@inproceedings{blip2,
  title={Blip-2: Bootstrapping language-image pre-training with frozen image encoders and large language models},
  author={Li, Junnan and Li, Dongxu and Savarese, Silvio and Hoi, Steven},
  booktitle={International conference on machine learning},
  pages={19730--19742},
  year={2023},
  organization={PMLR}
}

@article{graphtranslator,
  title={GraphTranslator: Aligning Graph Model to Large Language Model for Open-ended Tasks},
  author={Zhang, Mengmei and Sun, Mingwei and Wang, Peng and Fan, Shen and Mo, Yanhu and Xu, Xiaoxiao and Liu, Hong and Yang, Cheng and Shi, Chuan},
  journal={arXiv preprint arXiv:2402.07197},
  year={2024}
}

@misc{li2025largelanguagemodelsincontext,
      title={Are Large Language Models In-Context Graph Learners?}, 
      author={Jintang Li and Ruofan Wu and Yuchang Zhu and Huizhe Zhang and Liang Chen and Zibin Zheng},
      year={2025},
      eprint={2502.13562},
      archivePrefix={arXiv},
      primaryClass={cs.LG},
      url={https://arxiv.org/abs/2502.13562}, 
}

@misc{promptgfm,
      title={LLM as GNN: Graph Vocabulary Learning for Text-Attributed Graph Foundation Models}, 
      author={Xi Zhu and Haochen Xue and Ziwei Zhao and Wujiang Xu and Jingyuan Huang and Minghao Guo and Qifan Wang and Kaixiong Zhou and Yongfeng Zhang},
      year={2025},
      eprint={2503.03313},
      archivePrefix={arXiv},
      primaryClass={cs.LG},
      url={https://arxiv.org/abs/2503.03313}, 
}

@misc{dyprag,
      title={Dynamic Parametric Retrieval Augmented Generation for Test-time Knowledge Enhancement}, 
      author={Yuqiao Tan and Shizhu He and Huanxuan Liao and Jun Zhao and Kang Liu},
      year={2025},
      eprint={2503.23895},
      archivePrefix={arXiv},
      primaryClass={cs.CL},
      url={https://arxiv.org/abs/2503.23895}, 
}

@article{taglas,
  title={TAGLAS: An atlas of text-attributed graph datasets in the era of large graph and language models},
  author={Feng, Jiarui and Liu, Hao and Kong, Lecheng and Zhu, Mingfang and Chen, Yixin and Zhang, Muhan},
  journal={arXiv preprint arXiv:2406.14683},
  year={2024}
}

@inproceedings{pytorch,
  title={PyTorch: An Imperative Style, High-Performance Deep Learning Library},
  author={Paszke, Adam and Gross, Sam and Massa, Francisco and Lerer, Adam and Bradbury, James and Chanan, Gregory and Killeen, Trevor and Lin, Zeming and Gimelshein, Natalia and Antiga, Luca and Desmaison, Alban and K{\"o}pf, Andreas and Yang, Edward and DeVito, Zachary and Raison, Martin and Tejani, Alykhan and Chilamkurthy, Sasank and Steiner, Benoit and Fang, Lu and Bai, Junjie and Chintala, Soumith},
  booktitle={Advances in Neural Information Processing Systems},
  pages={8024--8035},
  year={2019}
}

@misc{huggingface_transformers,
      title={HuggingFace's Transformers: State-of-the-art Natural Language Processing}, 
      author={Thomas Wolf and Lysandre Debut and Victor Sanh and Julien Chaumond and Clement Delangue and Anthony Moi and Pierric Cistac and Tim Rault and Rémi Louf and Morgan Funtowicz and Joe Davison and Sam Shleifer and Patrick von Platen and Clara Ma and Yacine Jernite and Julien Plu and Canwen Xu and Teven Le Scao and Sylvain Gugger and Mariama Drame and Quentin Lhoest and Alexander M. Rush},
      year={2020},
      eprint={1910.03771},
      archivePrefix={arXiv},
      primaryClass={cs.CL},
      url={https://arxiv.org/abs/1910.03771}, 
}

@inproceedings{vllm,
  title={Efficient Memory Management for Large Language Model Serving with PagedAttention},
  author={Woosuk Kwon and Zhuohan Li and Siyuan Zhuang and Ying Sheng and Lianmin Zheng and Cody Hao Yu and Joseph E. Gonzalez and Hao Zhang and Ion Stoica},
  booktitle={Proceedings of the ACM SIGOPS 29th Symposium on Operating Systems Principles},
  year={2023}
}

@article{llama3,
  title={The llama 3 herd of models},
  author={Grattafiori, Aaron and Dubey, Abhimanyu and Jauhri, Abhinav and Pandey, Abhinav and Kadian, Abhishek and Al-Dahle, Ahmad and Letman, Aiesha and Mathur, Akhil and Schelten, Alan and Vaughan, Alex and others},
  journal={arXiv preprint arXiv:2407.21783},
  year={2024}
}

@article{codex,
  title={Codex: A comprehensive knowledge graph completion benchmark},
  author={Safavi, Tara and Koutra, Danai},
  journal={arXiv preprint arXiv:2009.07810},
  year={2020}
}

@article{nell23k,
  title={Dynamic anticipation and completion for multi-hop reasoning over sparse knowledge graph},
  author={Lv, Xin and Han, Xu and Hou, Lei and Li, Juanzi and Liu, Zhiyuan and Zhang, Wei and Zhang, Yichi and Kong, Hao and Wu, Suhui},
  journal={arXiv preprint arXiv:2010.01899},
  year={2020}
}

@article{clegr,
  title={A Graph Talks, But Who's Listening? Rethinking Evaluations for Graph-Language Models},
  author={Petkar, Soham and Vempati, Anirudh and Sinha, Akshit and Kumarauguru, Ponnurangam and Agarwal, Chirag and others},
  journal={arXiv preprint arXiv:2508.20583},
  year={2025}
}

@article{text_to_lora,
  title={Text-to-LoRA: Instant Transformer Adaption},
  author={Charakorn, Rujikorn and Cetin, Edoardo and Tang, Yujin and Lange, Robert Tjarko},
  journal={arXiv preprint arXiv:2506.06105},
  year={2025}
}

@article{lorahub,
  title={Lorahub: Efficient cross-task generalization via dynamic lora composition},
  author={Huang, Chengsong and Liu, Qian and Lin, Bill Yuchen and Pang, Tianyu and Du, Chao and Lin, Min},
  journal={arXiv preprint arXiv:2307.13269},
  year={2023}
}

@article{knots,
  title={Model merging with svd to tie the knots},
  author={Stoica, George and Ramesh, Pratik and Ecsedi, Boglarka and Choshen, Leshem and Hoffman, Judy},
  journal={arXiv preprint arXiv:2410.19735},
  year={2024}
}

@inproceedings{carlson2010toward,
  title={Toward an architecture for never-ending language learning},
  author={Carlson, Andrew and Betteridge, Justin and Kisiel, Bryan and Settles, Burr and Hruschka, Estevam and Mitchell, Tom},
  booktitle={Proceedings of the AAAI conference on artificial intelligence},
  volume={24},
  number={1},
  pages={1306--1313},
  year={2010}
}

@article{glfusion,
  title={Model generalization on text attribute graphs: Principles with large language models},
  author={Wang, Haoyu and Liu, Shikun and Wei, Rongzhe and Li, Pan},
  journal={arXiv preprint arXiv:2502.11836},
  year={2025}
}

@article{gfmrag,
  title={GFM-RAG: Graph Foundation Model for Retrieval Augmented Generation},
  author={Luo, Linhao and Zhao, Zicheng and Haffari, Gholamreza and Phung, Dinh and Gong, Chen and Pan, Shirui},
  journal={NeurIPS 2025},
  year={2025}
}

\appendix
\begin{table*}[!t]
% \small
 % \setlength{\tabcolsep}{2.3pt}
 % \renewcommand{\arraystretch}{1}
    \caption{Dataset statistics.}
    \vspace{-10pt}
    \label{tab:dataset}
    \centering
    \begin{tabular}{l|cccccc}
        \toprule
        Dataset &Avg. \# Node &Avg. \# Edge &Avg. \#Node words &Avg. \# Edge words & \# G \\ \midrule
        FB15K237 & 14,541 & 310,116 & 20.1 & 8.4 & 1 \\
        WN18RR & 40,943 & 93,003 & 23.3 & 1.0 & 1  \\
        CoDEx-Medium & 17,050 & 206,205 & 2.2 & 1.9& 1 \\
        NELL23K & 22,925 & 36,358 & 3.7 & 4.5 & 1 \\
        Scene Graph & 19.13 & 68.44 & 20.1 & 9.8 & 100000 \\
        Clegr-reasoning & 26.54 & 28.32 & 23.82 & 21.65 &  400 \\
        \bottomrule
    \end{tabular}
% }
\end{table*}
\begin{table*}[!t]
    \caption{The hyperparameter settings for all experiments}
    \vspace{-10pt}
    \centering
    \label{tab:hyperparameters}
    \begin{tabular}{c|cccccc}
    \toprule

    hyperparameters & Scene Graph & FB15K237 & WN18RR & Clegr-reasoning & CoDEx-Medium & NELL23K\\
    \midrule
       lora\_r  &  16 & 24 & 24 &  16 & 4 & 4 \\
       lora\_$\alpha$ & 32 & 48 & 48 & 32 & 8 & 8 \\ 
       lora components & MLP & MLP & MLP & MLP & MLP & MLP \\
         $N_s$ & 50 & 6000 & 6000 & 20 & 6000 & 6000\\
         $N_r$ & 100 & 2000 & 2000 & 160 & 2000 & 2000\\
         $N_c$ & 50 & 8000 & 8000 & 20 & 8000 & 8000\\
         stage1 minimum epoch & 5 & 1 & 1 & 5 & 1 & 1 \\
         stage1 maximum epoch & 50 & 1 & 1 & 50 & 1 & 1\\
         stage2 minimum epoch & 5 & 1 & 1 & 5 & 1 & 1\\
         stage2 maximum epoch & 50 & 5 & 5 & 50 & 10(qwen)/5(llama) & 10(qwen)/5(llama) \\
         early stop loss threshold & 0.4 & 0.15 & 0.15 & 0.4 & 0.15 & 0.15\\
         gradient accumulation steps & full & 512 & 512 & full & 512 & 512\\
         learning rate & 1e-3 & 1e-3 & 1e-3 & 1e-3 & 1e-3 & 1e-3\\
         scheduler & linear & linear & linear & linear & linear & linear \\
         $\beta_1$ & 0.9 & 0.9 & 0.9 & 0.9 & 0.9 & 0.9 \\
         $\beta_2$ & 0.98 & 0.98 & 0.98& 0.98 & 0.98 & 0.98 \\
         $\epsilon$ & 1e-4 & 1e-4 & 1e-4  & 1e-4 & 1e-4 & 1e-4 \\
         maximum gradient norm & 1.0 & 1.0 & 1.0 & 1.0 & 1.0 & 1.0\\
    \bottomrule
    \end{tabular}
    \label{tab:my_label}
\end{table*}

\section{Implementation Details}
\label{app:implementation}

In this section, we provide our implementation details of \methodname. 

\subsection{Implementation setting}
\label{app:implementation_setting}
We implement \methodname\ based on PyTorch~\cite{pytorch}, the Hugging Face Transformers library~\cite{huggingface_transformers}, and LIFT~\cite{lift}. The code for reproducibility is provided in \url{https://github.com/JiaruiFeng/GRIP}.

For the LoRA adapter, we apply LoRA by default to the MLP modules of each LLM layer. For subgraph sampling in both the summarization and reasoning QA task generation, we first randomly select a root node from the entire graph. We then iteratively expand the subgraph by sampling neighbors at each hop. Specifically, at each hop, we sample up to three neighbors per node and perform this process for three hops. This procedure results in a subgraph containing at most 10 nodes. For efficient task generation, we leverage the vLLM inference framework~\cite{vllm}. For generated fine-tuning tasks, we use plain text for context and summarization tasks and adopt an instruction-tuning template on all QA tasks. Each fine-tuning sample is formatted using the \texttt{apply\_chat\_template} function from the Hugging Face Transformers library. The exact user prompt and answer templates for context memorization, summarization, and edge-level context QA all follow short fixed patterns (e.g.\ ``In context graph \{title\}, the node \{src\} is \{rel\} the node \{tgt\}.''); the full set is released with the code.

\subsection{Prompts Design}
\label{app:prompts}
We use a single LLM (Qwen2.5-7b) to generate all fine-tuning tasks. Due to space limitation, here we only present the prompt design of the summarization task as a representative example. The remaining prompts (rephrasing, node-level / edge-level context QA, and the four reasoning-QA variants: multi-hop, global, binary, and $k$-shot) follow the same template and are all released with the code at \url{https://github.com/JiaruiFeng/GRIP}.
\begin{tcolorbox}[boxsep=0mm,left=2.5mm,right=2.5mm]
    \textbf{Summary:} You are given several text snippets, which representing real-world facts. Your task is to generate one summary to
    rephrase all information while exploring potential relationships across the snippets. Specifically: \\
    -Be accurate and complete, preserving ALL factual details like color or properties and numeric details like coordinate. \\
    -rewrite the original context in a completely different way, tone, writing style, and logic to summarize the original context. \\
    -Using rewriting techniques, including reordering, exchanging active and passive sentences, or synonym replacement. \\
    -If one text snippet, only do the summarization and rephrase. \\
    -If multiple text snippets are provided, focus on the relations between different text sentences, like multi-entity connections. \\
    The text snippets are provided below: \\
    \{context\}\\
    Please answer in the following format: Summary: [summary]. \\
    Please DON'T output quotes.
    \end{tcolorbox}

\section{Experiment details}
\label{app:experiment}

\subsection{Dataset details}
\label{app:data_stats}
In this section, we describe the details of all datasets we used for evaluation. The statistics of all datasets can be found in Table~\ref{tab:dataset}. The raw data of Scene Graph, FB15k237, and WN18RR datasets are obtained from TAGLAS~\cite{taglas}. The Clegr-reasoning~\cite{clegr}, CoDEx-Medium~\cite{codex}, and NELL23K~\cite{nell23k} are obtained from its original source. 

\textbf{Scene Graph.} Scene Graphs is a graph question answering dataset on scene graphs. Each graph in SceneGraphs contains objects connected by the relationship between two objects. It contains 59,978/19,997/20,025 graph samples for the train/val/test sets. Due to the time limitation, we only evaluate all methods on the first 500 test samples. 

\textbf{Clegr-reasoning.} Clegr-reasoning is a graph question answering generated by a syntactic subway line graph. Each graph contains several subway lines with intersections. It contains 400 different graphs and each graph has 34 questions generated from 34 different templates. These question templates test the model on different parts of graph reasoning, including topology, path, aggregation, and filtering. An example question template is: Which station is adjacent to both \{Station\} and \{Station\}. 

\textbf{FB15K237} FB15K237 is a knowledge graph. The dataset contains 14,541 nodes and 310,116 relations. Nodes in the dataset are entities in the knowledge graph and edges represent the relation between two entities. 
It contains 237 different relation types. There are a total of 272,115/17,535/20,466 samples for train/val/test sets, respectively. For simplicity, in evaluation, we only evaluate the randomly sampled 10000 samples from the test set, and each sample is formalized as a 10-way classification task. That is, the question will ask the model to select the correct one relation from 10 candidate relations, instead of all 237 relations. 

\textbf{WN18RR}
WN18RR is another knowledge graph extracted from WordNet. It contains 40,943 nodes and 93,003 relations, where each node is an English word and each edge represents the relation between two words. It contains 11 different relation types and 86,835/3,034/3,134 samples for train/val/test sets, respectively. For WN18RR, we evaluate on the whole test set. 

\textbf{CoDEx-Medium}
CoDEx-Medium is a knowledge graph built from Wikipedia/Wikidata. It contains 17,050 nodes and 206,205 relations, where each node is a Wikipedia/Wikidata-aligned entity and each edge denotes a factual relation between two entities. The dataset includes 51 relation types, with 185,584/10,310/10,311 relations in the train/val/test sets, respectively. Following the FB15K237 setup, we formalize each test set sample as a 10-way classification task, while we evaluate on the entire test set.

\textbf{NELL23K}
NELL23K is another knowledge graph constructed from the Never-Ending Language Learning (NELL) project~\cite{carlson2010toward}. It contains 22,925 nodes and 36,358 relations. Each node corresponds to a NELL concept instance (e.g., concept:person, concept:city), and each edge denotes a typed relation between two concepts. The dataset includes 200 different relation types, with 25,445/4,961/4,952 relations in the train/val/test sets, respectively. Following the CoDEx-Medium protocol, we cast each test sample as a 10-way classification task and evaluate on the entire test set.

\subsection{Training process and hyperparameters}

All experiments can be conducted on a single NVIDIA A100-SXM4-80GB GPU. To accelerate the training and inference process, we use 4 GPUs in total and split all datasets into 4 subsets and process them independently on each GPU. In Table~\ref{tab:hyperparameters}, we provide hyperparameter settings for all datasets.

For all knowledge graph datasets, the graph used in every \methodname\ stage (context memorisation, summarisation, and QA generation) contains only the training edges; validation and test edges are removed before any subgraph sampling or QA construction.

\end{document}